# Pre-trained Language Models and Few-shot Learning for Medical Entity Extraction

Xiaokai Wang
Santa Clara University
Santa Clara, USA

Guiran Liu
San Francisco State University
San Francisco, USA

Binrong Zhu
San Francisco State University
San Francisco, USA

Jacky He
Cornell University
New York, USA

Hongye Zheng
The Chinese University of Hong Kong
Hong Kong, China

Hanlu Zhang*
Stevens Institute of Technology
Hoboken, USA

*Abstract-This study proposes a medical entity extraction method based on Transformer to enhance the information extraction capability of medical literature. Considering the professionalism and complexity of medical texts, we compare the performance of different pre-trained language models (BERT, BioBERT, PubMedBERT, ClinicalBERT) in medical entity extraction tasks. Experimental results show that PubMedBERT achieves the best performance (F1-score = 88.8%), indicating that a language model pre-trained on biomedical literature is more effective in the medical domain. In addition, we analyze the impact of different entity extraction methods (CRF, Span-based, Seq2Seq) and find that the Span-based approach performs best in medical entity extraction tasks (F1-score = 88.6%). It demonstrates superior accuracy in identifying entity boundaries. In low-resource scenarios, we further explore the application of Few-shot Learning in medical entity extraction. Experimental results show that even with only 10-shot training samples, the model achieves an F1-score of 79.1%, verifying the effectiveness of Few-shot Learning under limited data conditions. This study confirms that the combination of pre-trained language models and Few-shot Learning can enhance the accuracy of medical entity extraction. Future research can integrate knowledge graphs and active learning strategies to improve the model's generalization and stability, providing a more effective solution for medical NLP research.*

## I. Introduction

Medical entity extraction is a key application of Natural Language Processing (NLP) in healthcare. With the rapid growth of biomedical research, the volume of medical literature is increasing exponentially. Each day, thousands of papers are added to databases such as PubMed, Medline, and Embase. Researchers must extract valuable information from this vast amount of data to support medical research, clinical decision-making, and drug development. However, traditional manual literature screening and analysis are time-consuming and labor-intensive, making it difficult to meet the demand for efficient information retrieval in modern medical research [1]. Entity extraction technology provides strong support for medical data mining by automatically identifying structured information from unstructured text, such as disease-drug relationships, gene-phenotype associations, and clinical treatment plans [2]. In recent years, deep learning, particularly the Transformer architecture, has significantly advanced NLP, improving the performance of medical entity extraction tasks.

Traditional medical entity extraction methods mainly rely on rule-based techniques and classical machine learning models. Rule-based approaches analyze text using predefined regular expressions, knowledge base matching, and expert-defined grammar rules. While these methods achieve high accuracy in specific tasks, their generalization ability is limited, making them ineffective for handling complex syntactic structures and diverse language expressions in medical literature. However, medical texts contain highly specialized terminology and hierarchical semantic structures. Traditional models often struggle with feature selection in large-scale literature and fail to capture deep contextual information. Efficient and accurate medical entity extraction has therefore become a central research challenge in medical NLP [3].

The introduction of Transformer-based models offers a new solution for medical entity extraction. Pre-trained language models such as BERT, RoBERTa, BioBERT, and PubMedBERT leverage self-attention mechanisms to learn long-range dependencies and capture deep semantic representations in medical texts. Beyond NLP, Transformer-based and hybrid architectures have also demonstrated strong performance in other domains. In computer vision and medical imaging, they have been applied to tasks such as 3D spine segmentation [4], skin disease detection [5], and object detection in clinical scans [6], showing enhanced accuracy through attention mechanisms and multi-scale fusion. In the field of multimodal learning, Transformer-CNN architectures have enabled more effective image-text classification through cross-modal feature fusion [7]. Additionally, in human-computer interaction, Transformer-related models have been used for optimizing interface design [8] and improving user experience through graph-based learning and dynamic adaptation [9]. These successes highlight the adaptability and effectiveness of Transformer models across diverse application areas [10-11]. Transformer-based models, such as BioBERT, have significantly improved medical entity recognition, relationship extraction, and text classification tasks. Fine-tuning and transfer learning enhance model adaptability, enabling better understanding of complex medical language structures compared to traditional methods [12].

This study proposes a Transformer-based framework to automate and enhance accuracy in medical entity extraction. Effective extraction supports rapid identification of medical evidence, aids clinical decision-making, and accelerates biomedical discoveries. Transformer models also enable multi-task learning and generative applications, including summarization and medical reasoning, thereby broadening their utility in medical NLP research [13]. As medical data expands, Transformer models will become increasingly important in precision medicine, diagnostics, and education. However, privacy and ethical issues require ongoing attention. Integrating multi-modal medical data and knowledge graphs with Transformer models may further enhance interpretability and scalability. This research aims to advance medical NLP by improving methods for processing extensive medical literature.

## II. RELATED WORK

The development of medical entity extraction has benefited greatly from advances in Transformer-based models and deep learning architectures. Notably, recent works have leveraged hierarchical and multimodal Transformer models to enhance named entity recognition (NER) performance. Tong et al. [14] proposed a semantic fusion framework using hierarchical Transformers, which enables the integration of diverse contextual representations. Similarly, prompt-based strategies have been introduced to optimize large language models for specialized entity extraction tasks, achieving better adaptability to domain-specific terminology and limited data scenarios [15].

Several studies have applied Transformer mechanisms within hybrid or multi-scale frameworks to strengthen deep representation learning. For instance, Hao et al. [16] designed a hybrid convolutional and Transformer-based architecture that captures both local and global dependencies in sequential data, improving the generalization of deep learning models. Multi-scale Transformer models further enhance performance by capturing hierarchical information, enabling more nuanced feature extraction [17]. Attention mechanisms, embedded within optimized neural architectures, also contribute to improved semantic segmentation and context-aware classification [18].

Beyond Transformer-based models, other neural network enhancements have significantly contributed to the advancement of deep feature extraction. Graph neural frameworks, particularly those that utilize self-supervised learning techniques, have been proposed to enhance the representation capabilities of complex data environments [19]. Additionally, dynamic rule mining mechanisms based on Transformer variants have been developed to facilitate adaptive pattern recognition in unstructured data [20].

Sequential modeling techniques also contribute valuable insights. LSTM-based prediction models have demonstrated robustness in handling time-dependent data and adaptive scheduling, providing lessons in efficient learning from limited sequences [21]. Combined with pattern discovery methods, such frameworks support enhanced spatiotemporal learning [22]. Additionally, deep neural network architectures have been effectively used to develop robust predictive systems, emphasizing the importance of carefully designed learning structures in handling heterogeneous and high-dimensional data [23].

## III. METHOD

This study proposes a medical literature information extraction model based on Transformer structure, which aims to automatically extract key entities and their relationships from medical texts. The self-attention mechanism architecture in Transformer is shown in Figure 1.

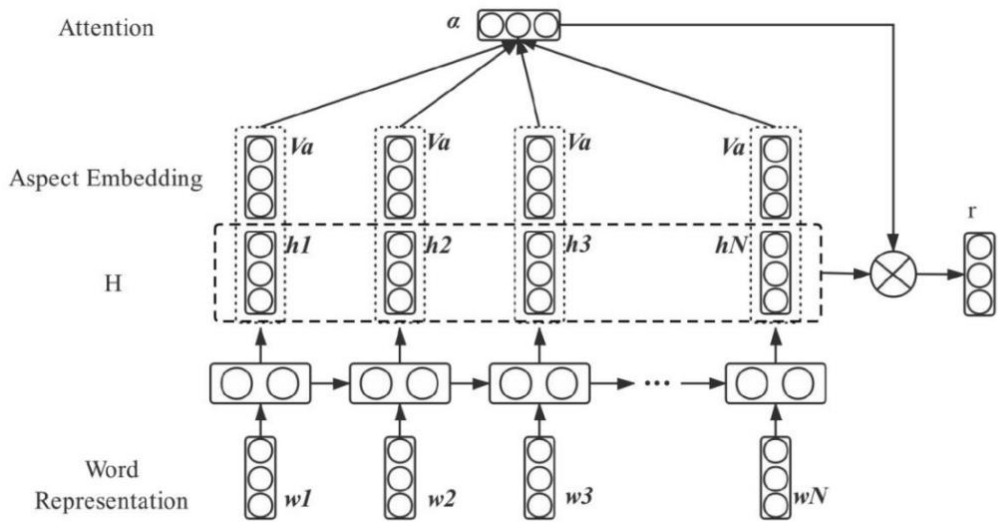

Figure 1. Basic structure diagram of self-attention mechanism

Given a medical text $X = \{x_1, x_2, ..., x_n\}$, where $x_i$ represents the i-th word, we first map the text to a high-dimensional vector space using the pre-trained medical domain BERT (such as BioBERT or PubMedBERT) to get the context representation $H = \{h_1, h_2, ..., h_n\}$ for each word. Transformer computes the relationships between words using a self-attention mechanism. The core calculation is as follows:

$$Attention(Q, K, V) = \text{softmax}(\frac{QK^T}{\sqrt{d_k}})V$$

Where $Q, K, V$ is the query matrix, the key matrix and the value matrix respectively, and $d_k$ is the scaling factor to stabilize the gradient update. Through multi-layer Transformer calculation, deep representation H of medical text can be obtained for subsequent information extraction tasks.

In the Medical named Entity recognition (NER) task [24], we use sequence annotation to feed Transformer's output H into a Conditional Random Field (CRF) layer to capture medical entity dependencies [25]. For a given label sequence $Y = \{y_1, y_2, ..., y_n\}$, define the conditional probability:

$$P(Y | X) = \frac{\exp(\sum_{i=1}^{n} Wy_{i-1}, y_i + h_i^T Wy_i)}{\sum_{Y'} \exp(\sum_{i=1}^{n} Wy'_{i-1} y'_i + h_i^T Wy'_i)}$$

Where W is the state transition matrix that controls dependencies between labels. The loss function is optimized with Negative Log-Likelihood (NLL):

$$L_{NER} = -\sum_{i=1}^{n} \log P(Y | X)$$

In relation extraction task, we adopt two-channel Transformer structure to independently model entity to $(e_1, e_2)$. First, we compute for each entity its context representation:

$$h_{e_i} = \frac{1}{|e_i|} \sum_{x_j \in e_i} h_j$$

Then, the entity pair representation is concatenated and the entity relationship score is calculated through a fully connected layer:

$$r_{e_1, e_2} = W_r [h_{e_1}][h_{e_2}] + b_r$$

Finally, cross-entropy loss is used to optimize relational classification:

$$L_{RE} = -\sum_{(e_1, e_2)} y_{e1, e2} \log P(r_{e_1, e_2})$$

Where, $y_{e1, e2}$ is the true label of the entity relationship, and $P(r_{e_1, e_2})$ is the probability of the relationship class predicted by the model. Through the joint optimization of NER and RE tasks, the entity recognition and relationship extraction promote each other and improve the accuracy of medical literature information extraction.

## IV. EXPERIMENT

### A. Datasets

This dataset comprises 6,881 disease entities extracted from PubMed abstracts, all of which were validated by biomedical experts and categorized into four groups—Specific, Composite, Modifier, and Undetermined Diseases. The data is split into training (5,064 instances), validation (787 instances), and test (1,030 instances) sets. A BIO tagging scheme (B-Begin, I-Inside, O-Outside) is used to clearly delineate entity boundaries, and all entities are aligned with the Unified Medical Language System (UMLS). Preprocessing steps include tokenization, stop-word removal, normalization, and Word Piece tokenization for medical terms. To address data imbalance, a Disease Co-occurrence Network was employed alongside data augmentation techniques such as synonym substitution and entity masking. These methods collectively bolster model performance in medical entity extraction tasks.

### B. Experimental Results

First, this paper gives the comparative experimental results of different pre-training models, as shown in Table 1.

Table 1. Performance comparison of different pre-trained language models on medical literature information extraction

| Model | Precision | Recall | F1-Score |
|---|---|---|---|
| Bert | 85.2% | 82.7% | 83.9% |
| BioBert | 88.4% | 86.1% | 87.2% |
| PubmedBert | 89.7% | 87.9% | 88.8% |
| ClinicalBert | 87.9% | 85.6% | 86.7% |

Experimental results indicate significant differences in the performance of various pre-trained language models in medical entity extraction tasks [26]. Among them, PubMedBERT achieves the highest performance across all evaluation metrics (F1-score = 88.8%). This suggests that its pre-training strategy on large-scale biomedical literature enhances its adaptability to the textual characteristics of medical texts. In comparison, BioBERT also demonstrates high accuracy (88.4%) and recall (86.1%), ranking second only to PubMedBERT. This indicates that BioBERT maintains strong generalization ability in specific medical entity extraction tasks. ClinicalBERT performs slightly worse than BioBERT and PubMedBERT, with a relatively lower recall (85.6%) despite achieving high accuracy (87.9%). This may be attributed to its pre-training on electronic health records (EHRs), which differ in textual style and structure from medical literature.

General BERT exhibits the weakest performance, with an F1-score of only 83.9%. This result highlights the limitations of general-purpose pre-trained language models when processing specialized medical texts. BERT is pre-trained on a general corpus and lacks domain-specific terminology and contextual understanding, making it less effective for medical entity extraction. In contrast, BioBERT and PubMedBERT, pre-trained on PubMed literature, improve their comprehension of medical terms, resulting in superior performance. PubMedBERT's advantage over BioBERT may stem from its training approach. While BioBERT fine-tunes BERT on biomedical texts, PubMedBERT is trained from scratch on medical literature. This allows PubMedBERT to capture the linguistic distribution and structural patterns of biomedical texts more comprehensively.

The findings indicate that choosing domain-specific pre-trained language models substantially enhances performance in medical entity extraction tasks. PubMedBERT and BioBERT outperform other models in both accuracy and recall. This suggests that generic BERT alone is inadequate for medical Natural Language Processing (NLP) tasks, and domain-adaptive pre-training strategies are crucial for improving model performance. Future research could delve deeper into integrating pre-trained language models with knowledge graph augmentation or multi-task learning to further enhance medical entity extraction capabilities.

Further, this paper also provides experimental comparative analysis of different named entity recognition methods based on Transformer, and the experimental results are shown in Table 2.

Table 2. Performance comparison of different named entity recognition methods based on Transformer

| Method | Precision | Recall | F1-Score |
|---|---|---|---|
| Transformer + CRF | 88.1% | 86.3% | 87.2% |
| Transformer + Span-based | 89.4% | 87.8% | 88.6% |
| Transformer + Seq2Seq | 86.7% | 85.2% | 85.9% |

Experimental results indicate that different entity extraction methods exhibit varying performance within Transformer-based architectures. Among them, the Span-based approach achieves the highest F1-score (88.6%), demonstrating its superior ability to accurately identify entity boundaries in medical texts. Compared to CRF-based token-by-token sequential labeling, the Span-based method directly predicts entity boundaries, making it more effective in extracting complex medical terms and multi-word expressions.

The Transformer + CRF method scores slightly lower (87.2%) but excels at capturing dependencies between entities, making it suitable for structured medical texts. Conversely, the Transformer + Seq2Seq method yields a lower F1-score (85.9%), likely due to the decoder's errors in boundary recognition during text generation. These findings suggest that Span-based and CRF methods are more effective for medical entity extraction. Future studies might explore integrating Span-based techniques with CRF to enhance accuracy and stability. Additionally, the study explores Few-shot Learning for medical entity recognition under low-resource conditions, with experimental results detailed in Figure 2.

Experimental results indicate that the performance of Few-shot Learning in medical entity extraction improves as the number of training samples increases. When training data is extremely limited (1-shot or 5-shot), the model's precision, recall, and F1-score remain low, with the F1-score ranging between 60% and 72%. This suggests that the model struggles to accurately identify medical entities under severe data scarcity. The primary challenge lies in the abundance of specialized terminology in medical texts, making it difficult for the model to learn effective patterns from minimal data. However, with 10-shot training, performance improves significantly, with the F1-score reaching 79.1%. This result demonstrates that even a small increase in labeled data can substantially enhance the model's learning ability. This aligns with the fundamental characteristic of Few-shot Learning, which efficiently leverages limited samples.

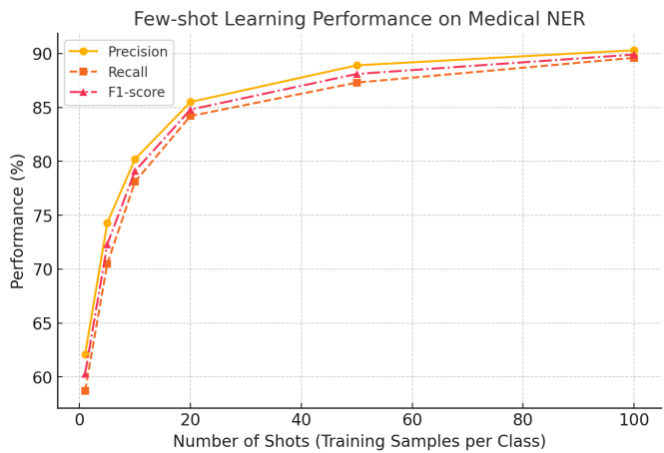

Figure 2. Few-shot Learning Performance on Medical NER

As the number of training samples increases to 20-shot and beyond, the model's precision, recall, and F1-score continue to rise, eventually plateauing after 50-shot. At 50-shot, the F1-score reaches 88.1% and approaches 89.9% at 100-shot. This indicates that with a sufficient number of training samples, the model can effectively learn entity representations and achieve high recognition accuracy. However, performance gains slow after 50-shot, suggesting that while Few-shot Learning enhances model performance up to a certain threshold, its marginal benefit diminishes as data volume increases. These findings suggest that in medical entity extraction tasks, Few-shot Learning is particularly effective in low-resource settings, whereas traditional supervised learning may offer greater stability when ample labeled data is available.

Overall, this experiment confirms the effectiveness of Few-shot Learning in medical entity extraction, particularly when labeled data is scarce. Even with a limited number of samples, the model demonstrates significant performance improvements. Future research could explore more advanced Few-shot Learning techniques, such as metric learning-based methods, GPT variants optimized for prompt design, or small-sample learning approaches integrated with knowledge graphs. These strategies could further enhance model generalization, enabling more effective applications in complex medical NLP tasks.

## V. CONCLUSION

This study proposes a medical entity extraction method based on Transformer and examines the effects of different pre-trained language models, extraction methods, and Few-shot Learning in low-resource scenarios. Experimental results indicate that PubMedBERT and BioBERT outperform other models in medical text processing, significantly improving entity extraction accuracy. Compared to traditional sequence labeling approaches, the Span-based entity extraction method achieves the best performance, demonstrating that directly predicting entity boundaries enhances the recognition of complex medical terms. Additionally, Few-shot Learning exhibits strong adaptability in low-resource conditions, achieving high F1-scores with minimal training data. This highlights its potential for medical NLP applications.

Despite these promising results, several aspects require further optimization. While the Transformer architecture enhances medical entity extraction, it incurs high computational costs, particularly on large-scale datasets. Future research could explore Knowledge Distillation or Lightweight Transformer variants to improve computational efficiency. This indicates that integrating Active Learning or Data Augmentation strategies may enhance model performance more efficiently. Additionally, medical texts often contain complex contextual relationships. Incorporating Knowledge Graphs into Transformer-based models could further strengthen their understanding of medical terminology.

Future studies could expand the application of Few-shot Learning in medical NLP, such as developing more effective prompt-based learning techniques. This would enable large language models (LLMs) to achieve high-precision entity extraction with minimal labeled data. Furthermore, real-world medical text data often involves privacy concerns. Optimizing medical entity extraction models while ensuring data security remains a critical research challenge. Finally, by refining Transformer architectures, integrating external medical knowledge, and introducing adaptive learning strategies, medical entity extraction technology could play a more significant role in clinical medicine, drug discovery, and medical literature analysis. These advancements would provide intelligent and efficient tools to support medical research and practice.